# On the Realization and Analysis of Circular Harmonic Transforms for Feature Detection

Hugh L. Kennedy

DST Group, ISS Division, Edinburgh, SA 5111, Australia
`Hugh.Kennedy@dst.defence.gov.au`

*Abstract*—Circular-harmonic spectra are a compact representation of local image features in two dimensions. It is well known that the computational complexity of such transforms is greatly reduced when polar separability is exploited in steerable filter-banks. Further simplifications are possible when Cartesian separability is incorporated using the radial apodization (i.e. weight, window, or taper) described here, as a consequence of the Laguerre/Hermite correspondence over polar/Cartesian coordinates. The chosen form also mitigates undesirable discretization artefacts due to angular aliasing. The possible utility of circular-harmonic spectra for the description of simple features is illustrated using real data from an airborne electro-optic sensor. The spectrum is deployed in a test-statistic to detect and characterize corners of arbitrary angle and orientation (i.e. wedges). The test-statistic considers uncertainty due to finite sampling and clutter/noise.

I. INTRODUCTION

The detection, classification, and matching, of simple geometric features are critical low-level operations in image-registration and target-tracking functions of modern image/video-processing systems [1],[2],[3],[4],[5]. Cartesian coordinates $(x, y)$, are convenient for digital image storage, display and filtering; however, local polar coordinates $(r, \theta)$, are ideal for image analysis and the rotation-invariant description of primitive shape information. First- and second-order Cartesian derivatives or bivariate quadratic polynomials are an adequate representation of simple features such as edges, ridges, peaks, and right-angle corners [2],[3],[6],[7],[8]. However, more complex features such as junctions (e.g. street intersections in satellite imagery) and wedges (e.g. ship wakes, aircraft wings or the corners of non-rectangular buildings) have simpler representations in polar coordinates [9],[10],[11],[12]; particularly when the dependence of the intensity on the radial and angular coordinates is assumed to be separable and the radial dependence is assumed to be constant over the chosen spatial scale.

This angle-only description of a known feature of interest (i.e. a *template*) or an unknown *patch* of (real) pixels is remarkably simple yet effective. A linear combination of circular harmonics $\varphi_l(\theta) = e^{il\theta}$ (where $i$ is the imaginary unit and $l$ is the angular wavenumber), captures the salient characteristics of the local angular profile, and the fidelity of the representation increases with the number of harmonic components considered; furthermore, the profile is readily rotated through an arbitrary angle by shifting the phase of each component appropriately. The vector $\boldsymbol{c}$ of complex coefficients $c_l$ used in the linear combination (i.e. the angular spectrum) is a compact feature descriptor. The descriptors of a patch and a template, or multiple patches in consecutive image frames, may then be compared to form a test statistic to support detection/classification or matching decisions. As an alternative to the use of so-called 'hand-crafted' features considered here, there is growing interest in the incorporation of rotation invariance – realized via "atomic filters" [13], "orientation channels" [14], "basis filters" [15], "steerable representations" [16], or "scale steerable filters" [17], for instance – in convolutional neural-networks.

In this note, the image is processed using a sequence of convolution operations realized in the pixel domain via a bank of linear basis-filters with a finite impulse-response (FIR) of square ($M \times M$) support in two dimensions (2-D). Moreover, a (complex) linear combination of their (complex) outputs may be formed to synthesize the output of a virtual filter with a response that is steered in an arbitrary direction. The impulse response of the $l$th basis filter $h_l(m_x, m_y)$, with frequency response $H_l(\omega_x, \omega_y)$, is simply defined by sampling the $l$th basis function $\psi_l(r, \theta)$, on a uniform Cartesian grid.

The general framework discussed above is often described in the literature, although there are substantial differences in the details. One of the most obvious points of departure is the way in which the integration/summation over the radial coordinate is weighted, shaped, or tapered (i.e. apodized) for scale selectivity and the suppression of spurious artefacts caused by window truncation and image discretization. Various approaches for polar-separable bases are surveyed in Section II.A. Laguerre-function components are used here as the radial weight because Cartesian-separable realizations are reached [18],[19],[20], with $2M \times (l + 1)$ multiply and add operations per basis-filter per pixel instead of $M^2$ (for pixel-domain realizations). This approach is detailed in Section III.A. Various ways of testing detection, classification, and matching, hypotheses using the angular spectrum are also surveyed in Section II.B. A test statistic for wedge detection is then described in Section III.B. In Section IV it is compared with some other simple yet popular approaches using simulated (see Section IV.A) and real (see Section IV.B) data. The test statistic and example application are provided to motivate and illustrate the utility and flexibility of circular-harmonic transforms in an image/video-processing context.

## II. Context

*A. Filter realization*

Bessel functions are used to represent radial oscillation in the polar Fourier transform; they are also radial eigenfunctions of the polar Laplacian [21]. Indeed, any family of orthogonal functions (e.g. Laguerre-Gauss [22],[23],[24],[25]) may in principle be used to represent radial form; however, a constant is sufficient for the primitive shapes considered here. The radial dependence $w(r)$, of the FIR filters is instead designed to set the scale of analysis $\lambda$ (in pixels), which is particularly important in multiscale methods, and to maintain the fidelity of the angular transform. Polar transforms are defined in the continuous domain and it is usually assumed that the required orthonormality of the basis-functions is retained in the discrete Cartesian coordinates of the digitized sensor data [10]; however, this is not always the case.

For 2-D FIR filters, rotational invariance and magnitude isotropy are degraded if the ideal impulse response is prematurely truncated by the finite window of the filter. Furthermore, the reproduction fidelity of high angular-frequencies on a Cartesian-grid diminishes as $r = 0$ is approached. This phenomenon is quantified in [9], using an angular analog of the Nyquist sampling theorem. It is also acknowledged in [26]; however, the form of the radial-weight applied is not specified there. The Laplacian of Gaussian [12], radial Gaussians [13], log radial (scaled by a Gaussian divided by a radial power) [17], triangular window [27], and various trigonometric forms on the unit disk [28], are examples of radial profiles defined in the spatial domain. The first scale of the Meyer-type profile [12], Simoncelli's bio-inspired isotropic wavelet [9], Erlang functions [29], and Log-Gabor responses with an infinite number of vanishing moments for multiscale analysis [30],[31], are examples of radial profiles defined in the frequency domain. A numerical optimization procedure is used to maximally localize wavelet frames, using measures of variance, in either the spatial or transform domains in [32]. Responses are bandlimited and polar-separable in all cases. Despite their apparent diversity, all rotationally-invariant image-analysis methods, e.g. multiscale circular-harmonics [24],[25], Fourier histograms of oriented gradients [5],[27], steerable wavelets [9],[12],[32] and high-order Riesz transforms [30], are fundamentally similar [10],[33]. All methods that use circular-harmonic expansions to describe the shape of local image features may potentially benefit from the simplification (i.e. Cartesian-separable realizations) described in this note.

It is well known that steerable filters derived from Hermite functions (or derivatives of a Gaussian) are separable in Cartesian coordinates; indeed, low-order derivatives (i.e. 0th, 1st and 2nd) are sufficient in many applications [2],[3],[6],[7],[8]. Such filtering operations are sometimes approximated and accelerated using recursively-realized box-kernels [1]. Closed-form expressions for the interconversion of Hermite functions (in Cartesian coordinates) and Laguerre-Gauss functions (in polar coordinates) have also been derived [22],[23]; however in these treatments, computational efficiency is not the aim and Cartesian separability is not exploited. For instance: in [22], Cartesian (astronomical) sensor measurements are converted to polar coordinates to support the rotational invariant analysis of spiral galaxies; and in [23], polar (biomedical) sensor measurements are converted to Cartesian coordinates for display and visualization. Furthermore in [24], it is shown that a Laguerre-Gauss function may be decomposed into its component terms and that it is sufficient to work with these non-orthonormal components; however, the 2-D FIR kernels of the corresponding basis filters in Cartesian coordinates are not decoupled and separated into perpendicular 1-D kernels. In [34], it is stated that Laguerre-Gauss circular-harmonics yield non-separable filters and that processing an entire image is time consuming. However, in Section III.A of this note (and in [18],[19],[20]) it is shown that Cartesian-separable realizations are reached if radial terms, referred to here as Laguerre-Gauss components $\mathcal{L}_l(r)$, are used as the radial weight $w_l(r)$, for the corresponding angular terms $\varphi_l(\theta)$. The resulting polar-separable basis-filters $\psi_l(r,\theta) = w_l(r)\varphi_l(\theta)$, have the required behavior for large $r$ (due to the Gaussian decay) and for small $r$ (due to the monomial notch).

*B. Spectrum analysis*

It is well known that the matched filter is an optimal detector that maximizes the signal-to-noise power ratio for a given signal in white noise; however, its performance may be very unsatisfactory in the presence of structured noise (e.g. clutter, interference, or even a dc offset) [35]. The matched filter is a sliding inner product and it is used to correlate angular spectra in [12] for the estimation of junction orientation angle (location and detection are not considered). Normalization of the feature template and the image patch yields a dimensionless test-statistic on the $[-1,1]$ interval [26]; however, the detection probability ($P_D$) is then rendered independent of intensity, which elevates the probability of false alarm ($P_F$) on dim structured noise. These correlation metrics are measures of *similarity*.

The sum-of-squared residuals (i.e. the square of the error norm) for a patch that is regressed against a template is a measure of *dissimilarity* and it may be more robust in the presence of structured noise. It is also used to match multi-scale circular-harmonic spectra [30]. Other quadratic similarity and dissimilarity measures for matching angular spectra are analyzed in [36], and a sparsity-based method for texture recognition is used in [33]. Templates are defined by maximizing an overlap integral, subject to a normalization constraint in [9],[12]; this approach is a continuous generalization of the piecewise-constant method used to design Slepian windows [33].

The dimensionless ratio of a model coefficient and an error standard-deviation (or their squares), estimated from randomly sampled populations, is routinely used as a test statistic in linear regression analysis [37]. In the context of a binary classifier, a detection event is declared when the coefficient-is-zero hypothesis (i.e. the 'null' hypothesis of the test) is rejected. The probability of a false declaration (i.e. $P_F$), when the null hypothesis is indeed true (i.e. the 'size' of the test), is determined from the tails of Student's t-distribution (or the tail of Snedecor's F-distribution). This test statistic may be interpreted as a ratio of similarity to dissimilarity. A test statistic of this form is adopted in this note because it is posited to have a higher $P_D$ (i.e. the 'power' of the test) for a given $P_F$, in the presence of structured clutter, which is always present in images/videos of real-world scenes. A large denominator is an indication of a poor model, which means that a large numerator is less meaningful. Such metrics may be interpreted as a signal-to-noise ratio (SNR) [11]. In Section III.B an integral analog of this statistic is evaluated from the angular spectrum $c$, of an image patch in the transform domain, for a wedge template of a specified angular width $2\phi_\Delta$, in the spatial domain. The test statistic is evaluated for a finite number of possible wedge orientations $\theta_\Delta$.

## III. DETAILS

### A. Filter realization

The Laguerre-Gauss $\mathcal{L}_l$, and Hermite-function $\mathcal{H}_k$, components are defined here as

$$\mathcal{L}_l(r) = r^l e^{-r^2/2\lambda^2}$$

$$\mathcal{H}_k(x) = x^k e^{-x^2/2\lambda^2} \text{ and } \mathcal{H}_k(y) = y^k e^{-y^2/2\lambda^2}. \quad (1)$$

They are used below to facilitate the transformation of $\psi_l(r,\theta)$, into Cartesian-separable form. The polar basis-function $\psi_l^\mathcal{L}(r,\theta)$, or its Cartesian equivalent $\psi_l^\mathcal{H}(x,y)$, is sampled over the discrete Cartesian coordinates of the basis filter to yield $h_l^\mathcal{L}(m_x, m_y)$ where $m$ is an integer shift index (see Fig. 1). Using $w_l(r) = \mathcal{L}_l(r)$ and $\varphi_l(\theta) = e^{il\theta}$ yields

$$\psi_l^\mathcal{L}(r,\theta) = r^l e^{-r^2/2\lambda^2}[\cos(l\theta) + i\sin(l\theta)] \text{ or} \quad (2a)$$

$$\psi_l^\mathcal{L}(r,\theta) = e^{-r^2/2\lambda^2} \sum_{k=0}^{l} \Gamma_{k,l-k} r^k \cos^k(\theta) r^{l-k} \sin^{l-k}(\theta) \text{ where}$$

$$\Gamma_{k,l-k} = \binom{l}{k}\{\cos(\pi[l-k]/2) + i\sin(\pi[l-k]/2)\} \quad (2b)$$

after using the multiple-angle formulae. Substituting $r\cos\theta = x$, $r\sin\theta = y$ and $r^2 = x^2 + y^2$ into (2b) then yields

$$\psi_l^\mathcal{H}(x,y) = e^{-[x^2+y^2]/2\lambda^2} \sum_{k=0}^{l} \Gamma_{k,l-k} x^k y^{l-k} \text{ or} \quad (3a)$$

$$\psi_l^\mathcal{H}(x,y) = \sum_{k=0}^{l} \Gamma_{k,l-k} \mathcal{H}_k(x) \mathcal{H}_{l-k}(y). \quad (3b)$$

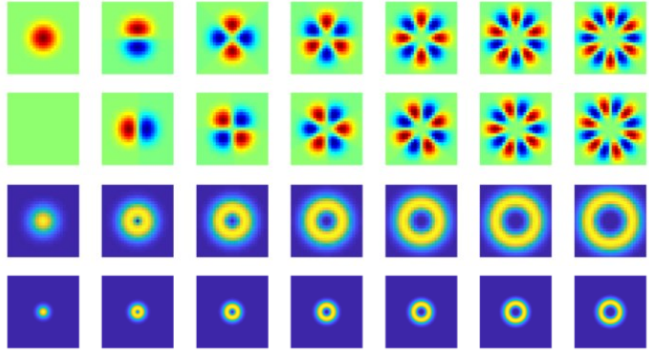

**Fig. 1**. Response of polar-separable basis-filters for $l = 0 \ldots 6$ (left to right). Real part, imaginary part, and magnitude, of impulse response $h_l^\mathcal{L}(m_x, m_y)$ over $m = \pm 12$, magnitude of frequency response $H_l^\mathcal{L}(\omega_x, \omega_y)$ over $\omega = \pm\pi$ (top to bottom). These basis-filters are not Cartesian separable but they may be formed from a sum of $l + 1$ component-filters that are (see Fig. 2).

Thus $\mathcal{H}_k$ is sampled over $-K \leq m \leq K$ (i.e. $M = 2K + 1$), and the indexing direction reversed, to produce the 1-D kernels of the (real) Hermite-function component-filters $h_k^\mathcal{H}(m)$ (see Fig. 2). The $l$th angular spectrum coefficient $C_l$ at pixel index $(n_x, n_y)$, for $0 \leq l \leq L$ and $0 \leq n < N$, is therefore evaluated using two consecutive 1-D convolutions followed by a (complex) linear combination:

$$A_{k_x}(n_x, n_y) = \sum_{m_x=-K}^{K} h_{k_x}^\mathcal{H}(m_x) I(n_x - m_x, n_y) \text{ and} \quad (4)$$

$$B_{k_x,k_y}(n_x, n_y) = \sum_{m_y=-K}^{K} h_{k_y}^\mathcal{H}(m_y) A_{k_x}(n_x, n_y - m_y) \text{ then}$$

$$C_l(n_x, n_y) = \rho_l^{-1/2} \sum_{k=0}^{l} \overline{\Gamma}_{k,l-k} B_{k,l-k}(n_x, n_y) \text{ where}$$

$$\rho_l = \sum_{m_x=-K}^{K} \sum_{m_y=-K}^{K} |h_l^\mathcal{L}(m_x, m_y)|^2 \text{ and}$$

$I(n_x, n_y)$ is the (real) input image ($\overline{[\cdot]}$ denotes complex conjugation).

An isotropic magnitude response in the continuous frequency- domain $(\omega_x, \omega_y)$, indicates that the basis-functions are faithfully reproduced and that the discretized basis-set is approximately orthogonal (see Fig. 1, bottom row). This is generally the case for a given scale (as set using $\lambda$) if $l$ is not too large and $K$ is not too small. Imperfections may be less noticeable in the discrete spatial-domain $(m_x, m_y)$.

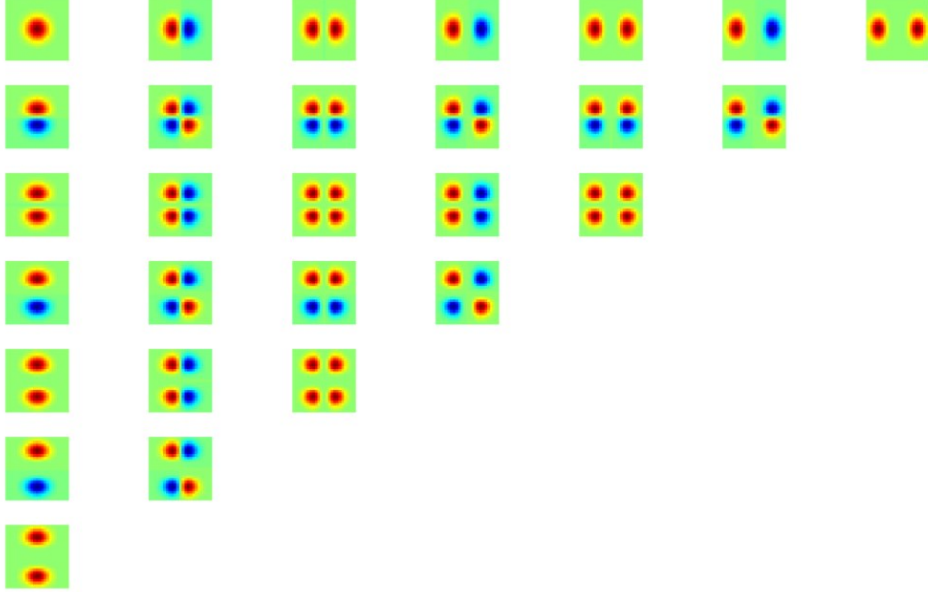

**Fig. 2**. Impulse responses of Cartesian-separable Hermite-function component-filters $h_{k_x}^{\mathcal{H}}(m_x) h_{k_y}^{\mathcal{H}}(m_y)$; $k_x = 0 \ldots 6$ (top to bottom); $k_y = 0 \ldots 6$ (left to right). Within each subplot: $m_x = -12 \ldots 12$ (top to bottom); $m_y = -12 \ldots 12$ (left to right). The $l$th basis-filter (see Fig. 1) is a complex linear combination (using coefficients $\Gamma_{k,l-k}$) of the component-filters along the $l$th diagonal, running from upper right to lower left.

*B. Spectrum analysis*

Negative wavenumbers are included using $C_{-l} = \bar{C}_l$ to make full use of the rotational symmetry afforded by a complex representation in the equations below. The angular spectrum $\boldsymbol{c}$ (a column vector of length 2$L$+1), is formed from $C_l(n_x, n_y)$ for $-L \leq l \leq L$ at a given pixel. It may then be used to reconstruct the angular dependence of the local image intensity via the inverse circular-harmonic transform

$$\hat{I}(\theta) = \sum_{l=-L}^{L} c_l e^{il\theta}. \qquad (5)$$

This inverse transform allows the circular-harmonic functions to interpolate in the angular coordinate. However, in addition to visual clutter and interference, sampling and truncating in the Cartesian domain yields spurious oscillatory artefacts in the locally-fitted angular-function. The test-statistic $Z_t$ presented below, incorporates and considers these imperfections, via a variance term, when the structure of a local image feature is analyzed. Fortunately, the required integrals of $\hat{I}(\theta)$ are readily evaluated in the transform domain using $\boldsymbol{c}$.

The mean and variance $(\mu, \sigma^2)$ of $\hat{I}(\theta)$, over the inner and outer domains of the wedge template (denoted using $[\cdot]_1$ and $[\cdot]_0$ subscripts, respectively) are computed for a given steering angle and used to evaluate $Z_t(n_x, n_y)$ as follows:

$$Z_t = (\mu_1 - \mu_0)/\sqrt{\sigma_1^2 + \sigma_0^2 + \sigma_{\min}^2} \qquad (6a)$$

where

$\mu_1 = \boldsymbol{s}_1^\intercal \boldsymbol{c}_\theta/(2\phi_1)$,
$\mu_0 = \boldsymbol{s}_0^\intercal \boldsymbol{c}_\theta/(2\pi - 2\phi_0)$
$\sigma_1^2 = \boldsymbol{c}_\theta^\dagger \boldsymbol{S}_1 \boldsymbol{c}_\theta/(2\phi_1) - \mu_1^2$
$\sigma_0^2 = \boldsymbol{c}_\theta^\dagger \boldsymbol{S}_0 \boldsymbol{c}_\theta/(2\pi - 2\phi_0) - \mu_0^2$
$\sigma_{\min}^2$ is the minimum intensity variance (a fixed parameter)
$\phi_1 = \phi_\Delta - \epsilon_\Delta$
$\phi_0 = \phi_\Delta + \epsilon_\Delta$

$2\phi_\Delta$ is the angular width of the wedge (the inner domain)

$2\epsilon_\Delta$ is the width of the gap between the inner and outer domains

$\mathbf{c}_\theta$ is the steered angular spectrum with elements $c_l e^{il\theta_\Delta}$

$\theta_\Delta$ is the hypothesized orientation angle of the wedge

$([\cdot]^\top$ is a transpose, $[\cdot]^\dagger$ is a Hermitian transpose). (6b)

The $l$th elements of $\mathbf{s}_1$ & $\mathbf{s}_0$ (column vectors) and the elements in the $l_m$th row and $l_n$th column of $\mathbf{S}_1$ & $\mathbf{S}_0$ (symmetric matrices) are, respectively:

$$\int_{-\phi_1}^{\phi_1} e^{il\theta} d\theta = \mathcal{S}_l(\phi_1)$$

$$\int_{-\pi}^{-\phi_0} e^{il\theta} d\theta + \int_{\phi_0}^{\pi} e^{il\theta} d\theta = \mathcal{S}_l(\pi) - \mathcal{S}_l(\phi_0)$$

$$\int_{-\phi_1}^{\phi_1} e^{i(l_m-l_n)\theta} d\theta = \mathcal{S}_{l_m-l_n}(\phi_1) \text{ and}$$

$$\int_{-\pi}^{-\phi_0} e^{i(l_m-l_n)\theta} d\theta + \int_{\phi_0}^{\pi} e^{i(l_m-l_n)\theta} d\theta = \mathcal{S}_{l_m-l_n}(\pi) - \mathcal{S}_{l_m-l_n}(\phi_0) \text{ where}$$

$$\mathcal{S}_l(\phi_c) = \begin{cases} 2\sin(l\phi_c)/l & l \neq 0 \\ 2\phi_c & l = 0 \end{cases}. \quad (7)$$

The $\mathcal{S}_l$ auxiliary functions are also used in Slepian designs [33]. All elements are real and precomputed offline. The test statistic is large: when the difference of intensity means is large, i.e. the numerator of (6a); and when the sum of intensity variances is small, i.e. the denominator of (6a); over the inner and outer domains of the wedge template. An arbitrary decision threshold is applied ($Z_t > \lambda_\Delta$). The $Z_t$ statistic is signed; therefore, it is used to discriminate between bright and dark objects; otherwise, $Z_F = Z_t^2$ is used. The estimate of the wedge's orientation $\hat{\theta}_\Delta$, is set equal to the angle $\theta_\Delta$, that maximizes $Z$. The joint estimation of $\theta_\Delta$ and $\phi_\Delta$ is not attempted here.

## IV. Application and Illustration

### A. Synthetic data

The Area Under Curve (AUC) of the Receiver Operating Characteristic (ROC) for the proposed wedge detector (Det. A), a detector with a maximally concentrated (i.e. Slepian) wedge template (Det. B), a least-squares fitted wedge template (Det. C), a Harris corner detector (Det. D, [4],[26],[38],[39]) and a Kitchen-Rosenfeld detector (Det. E, [39],[40]) were examined (see Table I). Dets. B & C both used a standard correlation-type measure of similarity (see Section II.B).

Each (25 x 25) synthetic frame contained a wedge with an intensity of 255 on a background with an intensity of 100. Other wedge parameters were randomly drawn from the following uniform distributions: $2\tilde{\phi}_\Delta \sim \mathcal{U}(\pi/12, \pi)$, $\tilde{\theta}_\Delta \sim \mathcal{U}(0,2\pi)$, $\tilde{r}_\Delta \sim \mathcal{U}(0,6)$, $\tilde{\vartheta}_\Delta \sim \mathcal{U}(0,2\pi)$. The displacement of the wedge apex from the center of the frame is $\Delta\tilde{x} = \tilde{r}_\Delta \cos\tilde{\vartheta}_\Delta$ and $\Delta\tilde{y} = \tilde{r}_\Delta \sin\tilde{\vartheta}_\Delta$. The test statistic ($Z_t$) at the center of the frame was evaluated. The achieved $P_F$ and $P_D$ for a range of threshold settings was computed, from 10,000 random instantiations of each angle scenario. For a given wedge width $2\phi_\Delta$ and threshold $\lambda_\Delta$ combination, any detection for a wedge with $\tilde{r}_\Delta < 2$ and $2\tilde{\phi}_\Delta > 2\phi_\Delta - \pi/12$ and $2\tilde{\phi}_\Delta < 2\phi_\Delta + \pi/12$ is deemed to be true; otherwise it is false. The $\theta_\Delta$ grid was quantized using steps of $\pi/12$ (for Dets. A-C); the accuracy of $\hat{\theta}_\Delta$ is not considered here. Two random instantiations and their corresponding $Z_t$ calculations are illustrated in Fig. 3.

Det. A has the highest AUC for all angles and detectors in all but one case. The AUC of Dets. D & E is maximized for right-angle corners as expected. Unlike the least-squares design of Det. C and the integral metric of Det. A, the eigen-design of Det. B works best for narrower wedges.

TABLE I
AUC[*] OF ROC FOR VARIOUS WEDGE WIDTHS AND DETECTORS

| $2\phi_\Delta$ (°) | A [a,b] | B [b] | C [b] | D | E |
|---|---|---|---|---|---|
| 45 | 0.9074 | 0.8739 | 0.8365 | 0.8385 | 0.7652 |
| 60 | 0.9133 | 0.9196 | 0.8519 | 0.9071 | 0.8262 |
| 90 | 0.9522 | 0.8816 | 0.8634 | 0.9099 | 0.8743 |
| 120 | 0.9409 | 0.7795 | 0.8486 | 0.7906 | 0.8348 |
| 135 | 0.9512 | 0.7064 | 0.8355 | 0.6781 | 0.7689 |

[*]A perfect detector has unity AUC
[a] $\epsilon_\Delta = \theta_\Delta/3$, $\sigma_{min}^2 = 255^2$; [b] $L = 6$, $K = 12$ and $\lambda = 3$ pix (see Fig. 1)

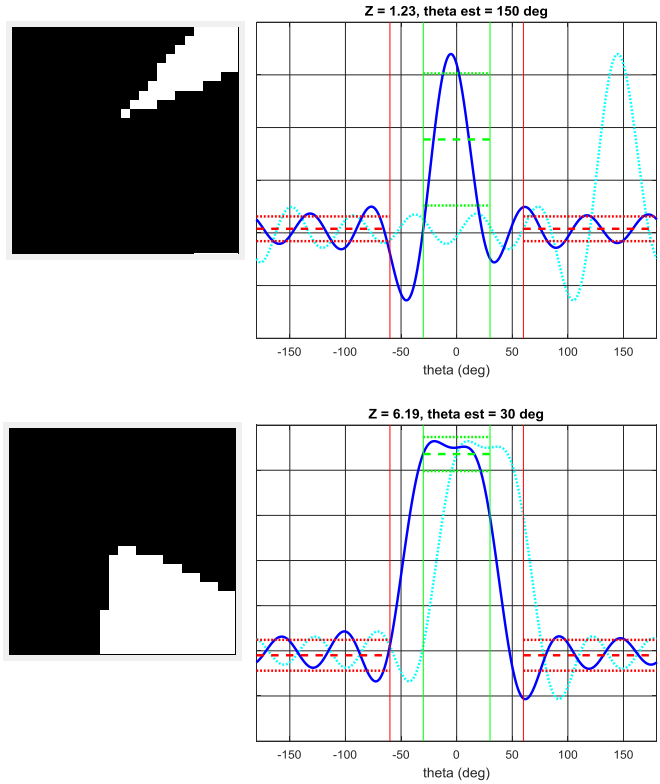

**Fig. 3**. Sample image instantiations for ROC simulations (left) and plots of $\hat{I}(\theta)$ vs. $\theta$ for test-statistic evaluations (right) for false (top) and true (bottom) detections for a template tuned to $2\phi_\Delta = 90°$. Dotted cyan: $\hat{I}(\theta)$ in image coordinates; Solid blue: $\hat{I}(\theta)$ in template coordinates (i.e. rotated through $-\hat{\theta}_\Delta$); Solid green: $\pm\phi_1$; Dashed green: $\mu_1$; Dotted green: $\mu_1 \pm \sigma_1$; Solid red: $\pm\phi_0$; Dashed red: $\mu_0$; Dotted red: $\mu_0 \pm \sigma_0$; $Z_t$ & $\hat{\theta}_\Delta$ evaluations are also shown.

*B. Real Data*

Chips of monochrome image sequences (1024 x 1024, 8-bit, @ 3 Hz) collected from a wide-area airborne-sensor over Port Adelaide were processed using various detectors. All detectors label most of the obvious corners in the scene (e.g. see Figs. 4 & 5, top 200 detections shown); however, the Harris corner-detector (Det. D) has a proclivity for the ends of lines and small blobs, the $Z_t$ wedge-detector (Det. A) for non-centered edges; and for this reason, the latter algorithm was tuned using $2\phi_\Delta = 60°$ (instead of 90°) to attenuate the edge response. When it is tuned using $2\phi_\Delta = 270°$, the warehouse in the yellow box (with a dark 90° corner) has the largest $Z_t$ in the scene (see yellow inset of Fig. 6). When the filter-bank is steered to $\theta_\Delta = -45°$ only, the missed corner on the warehouse in the green box is more prominent (see green inset of Fig. 6). Generation of the **c** spectrum at each pixel of this image takes approximately 1.3 s; then approximately 0.6 s per $\theta_\Delta$ hypothesis for the $Z_t$ statistic evaluation (using a MATLAB® script with no toolboxes running on a personal computer with an Intel® Core™ i7-6700HQ processor @ 3.4 GHz). Increasing $L$ generally improves angular selectivity by lowing angular side-lobes and decreasing the width of the inner/outer transition region; although as reported in [11] & [12], there are rapidly diminishing returns beyond $L = 6$ for this type of problem.

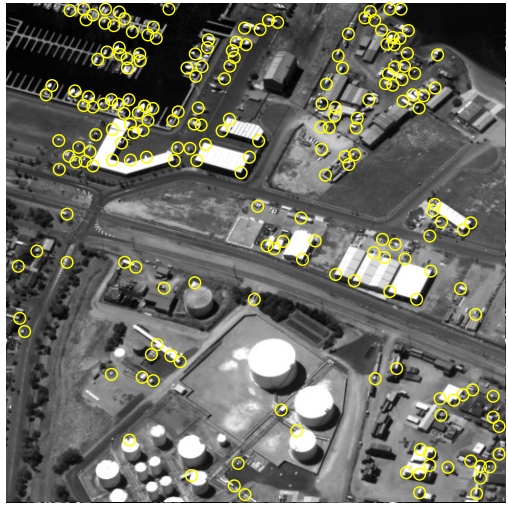

**Fig. 4**. Output of Harris corner-detector on the input image.

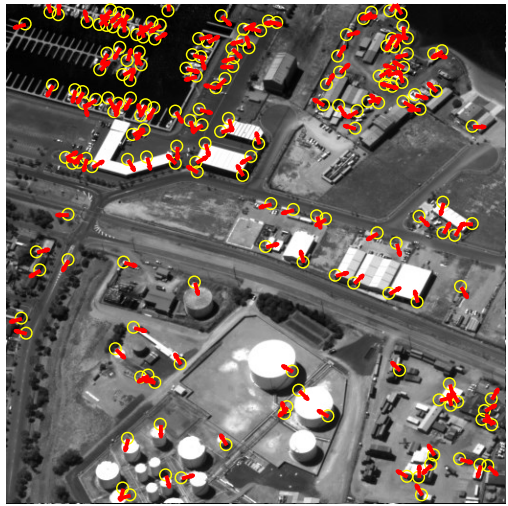

**Fig. 5**. Output of $Z_t$ wedge-detector, for $2\phi_\Delta = 60°$. Wedge locations in yellow and their estimated orientations ($\hat{\theta}_\Delta$) in red.

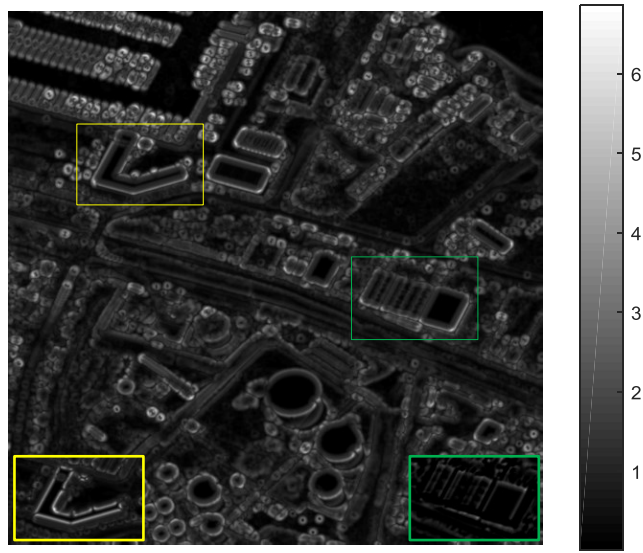

**Fig. 6**. Map of $Z_t$ for $2\phi_\Delta = 60°$; yellow inset contains detail for $2\phi_\Delta = 270°$, green inset contains detail for $\theta_\Delta = -45°$ only).

V. CONCLUSION

We live in a world where objects do not change as they are turned or viewed from a different perspective thus (continuous) polar or spherical coordinates are ideal for their representation. However, the digital machines we have built in recent times, to sense and perceive such scenes on our behalf, operate on a (discrete) Cartesian grid. It is therefore important to develop computer-vision systems with rotationally invariant properties that are able to reconcile this mismatch; for instance: using radially-weighted circular harmonics that are uniformly sampled and truncated in Cartesian coordinates.

In the computer-vision literature, there are a plethora of radial weights for steerable *polar*-separable basis-functions. In this note, the form of the radial weight is chosen for its ability to realize polar-separable responses in *Cartesian*-separable form for a significant reduction in the computational complexity of rotationally-invariant filter-banks. Factoring digital filtering operations in this way, allows 2-D convolutions to be replaced by consecutive 1-D convolutions for higher data-throughput rates in online image/video-processing systems. The radial weight also focuses the filter-bank around $r = 0$ (i.e. on the pixel-under test) while ensuring adequate angular resolution after discretization. The discrete spectrum of circular harmonics produced by these filters is used in an angular-integral test-statistic for the detection of wedges in images. Like traditional (t- and F-distributed) test statistics used in regression analysis, this similarity measure incorporates and considers uncertainty due to the limitations of finite sampling. Further benchmarking and performance comparisons, multiscale extensions, the introduction of a third (e.g. temporal) dimension, optimal weights for non-separable realizations, the joint estimation of feature width and orientation, and target tracking applications, will be investigated in future work.